\documentclass[11pt,a4paper]{article}
\usepackage[hyperref]{acl2019}
\usepackage{times}
\usepackage{latexsym}
\usepackage{xspace}
\usepackage{multirow}
\usepackage{url}
\usepackage{graphicx}
\usepackage{arydshln}   

\aclfinalcopy 

\newcommand\BLEU{\textsc{Bleu}\xspace}

\title{APE at Scale and its Implications on MT Evaluation Biases}

\author{Markus Freitag, Isaac Caswell, Scott Roy\\
  Google Research \\
  {\tt \{freitag,icaswell,hsr\}@google.com}}

\date{}

\begin{document}
\maketitle
\begin{abstract}
In this work, we train an Automatic Post-Editing (APE) model and use it to reveal biases in standard Machine Translation (MT) evaluation procedures.
The goal of our APE model is to correct typical errors introduced by the translation process, and convert the ``translationese" output into natural text.
Our APE model is trained entirely on monolingual data that has been round-trip translated through English, to mimic errors that are similar to the ones introduced by NMT.
We apply our model to the output of existing NMT systems, and demonstrate that, while the human-judged quality improves in all cases, \BLEU scores drop with forward-translated test sets. We verify these results for the WMT18 English$\to$German, WMT15 English$\to$French, and WMT16 English$\to$Romanian tasks. Furthermore, we selectively apply our APE model on the output of the top submissions of the most recent WMT evaluation campaigns. We see quality improvements on all tasks of up to 2.5 BLEU points.

\end{abstract}

\section{Introduction}
\label{sec:intro}
Neural Machine Translation (NMT) \cite{bahdanau2014neural, gehring2017convolutional, vaswani2017attention} is currently the most popular approach in Machine Translation leading to state-of-the-art performance for many tasks. NMT relies mainly on parallel training data, which can be an expensive and scarce resource. 
There are several approaches to leverage monolingual data for NMT:
Language model fusion for both phrase-based \cite{brants2007large} and neural MT \cite{gulcehre2015using,gulcehre2017integrating}, back-translation \cite{sennrich2016improving}, unsupervised NMT \cite{lample2017unsupervised, artetxe2018unsupervised}, dual learning  \cite{cheng2016semisupervised, he2016dual, xia2017dual}, and multi-task learning \cite{domhan2017using}.

In this paper, we present a different approach to leverage monolingual data, which can be used as a post-processor for any existing translation.
The idea is to train an Automatic Post-Editing (APE) system that is only trained on a large amount of synthetic data, to fix typical errors introduced by the translation process. 
During training, our model uses a noisy version of each sentence as input and learns how to reconstruct the original sentence.
In this work, we model the noise with
round-trip translations (RTT) through English, translating a sentence in the target language into English, then translating the result back into the original language.
We train our APE model with a standard transformer model on the WMT18 English$\to$German, WMT15 English$\to$French and WMT16 English$\to$Romanian monolingual News Crawl data and apply this model
on the output of NMT models that are either trained on all available bitext or trained on a combination of bitext and back-translated monolingual data.
Furthermore, we show that our APE model can be used as a post-processor for the best output of the recent WMT evaluation campaigns, where it improves even the 
output of these well engineered translation systems.

In addition to measuring quality in terms of \BLEU scores on the standard WMT test sets, we split each test set into two subsets based on whether the source or target is the original sentence (each sentence is either originally written in the source or target language and human-translated into the other). We call these the source-language-original and target-language-original halves, respectively. We find that evaluating our post-edited output on the source-language-original half actually decreases the \BLEU scores, whereas the \BLEU scores improve for the target-language-original half.
This is in line with results from \newcite{Koppel:2011:TD:2002472.2002636}, who demonstrate that the mere fact of being translated plays a crucial role in the makeup of a translated text, making the actual (human) translation a less natural example of the target language. We hypothesize that, given these findings, the consistent decreases in \BLEU scores on test sets whose source side are natural text does not mean that the actual output is of lower quality.
To verify this hypothesis, we run human evaluations for different outputs with and without APE.
The human ratings demonstrate that the output of the APE model is both consistently more accurate and consistently more fluent, regardless of whether the source or the target language is the original language, contradicting the corresponding \BLEU scores.

To summarize the contributions of the paper:
\begin{itemize}
    \item We introduce an APE model trained only on synthetic data generated with RTT for fixing typical translation errors from NMT output and investigate its scalability. To the best of our knowledge, this paper is the first to study the effect of an APE system trained at scale and only on synthetic data.
    \item We improve the \BLEU of top submissions of the recent WMT evaluation campaigns.
    \item We show that the \BLEU scores of the APE output only correlate well with human ratings when they are calculated with target-original references.
    \item We propose separately reporting scores on test sets whose source sentences are translated and whose target sentences are translated, and call for higher-quality test sets.
\end{itemize}

\section{APE with RTT}
\label{sec:ape_model}
\subsection{Definition and Training}
\begin{figure}[ht]
    \centering
    \def\svgwidth{\columnwidth}
\begingroup%
  \makeatletter%
  \providecommand\color[2][]{%
    \errmessage{(Inkscape) Color is used for the text in Inkscape, but the package 'color.sty' is not loaded}%
    \renewcommand\color[2][]{}%
  }%
  \providecommand\transparent[1]{%
    \errmessage{(Inkscape) Transparency is used (non-zero) for the text in Inkscape, but the package 'transparent.sty' is not loaded}%
    \renewcommand\transparent[1]{}%
  }%
  \providecommand\rotatebox[2]{#2}%
  \newcommand*\fsize{\dimexpr\f@size pt\relax}%
  \newcommand*\lineheight[1]{\fontsize{\fsize}{#1\fsize}\selectfont}%
  \ifx\svgwidth\undefined%
    \setlength{\unitlength}{204.79209271bp}%
    \ifx\svgscale\undefined%
      \relax%
    \else%
      \setlength{\unitlength}{\unitlength * \real{\svgscale}}%
    \fi%
  \else%
    \setlength{\unitlength}{\svgwidth}%
  \fi%
  \global\let\svgwidth\undefined%
  \global\let\svgscale\undefined%
  \makeatother%
  \begin{picture}(1,0.29163064)%
    \lineheight{1}%
    \setlength\tabcolsep{0pt}%
    \put(0,0){\includegraphics[width=\unitlength,page=1]{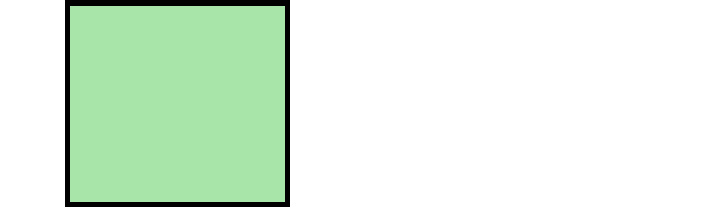}}%
    \put(0.12709192,0.18855027){\color[rgb]{0,0,0}\makebox(0,0)[lt]{\lineheight{1.25}\smash{\begin{tabular}[t]{l}\ RoundTrip\\Translations\\\ $Y$$\to$$X$$\to$$Y$\end{tabular}}}}%
    \put(0,0){\includegraphics[width=\unitlength,page=2]{repair_train.pdf}}%
    \put(0.68009226,0.15925224){\color[rgb]{0,0,0}\makebox(0,0)[lt]{\lineheight{1.25}\smash{\begin{tabular}[t]{l}\ APE\\Model\end{tabular}}}}%
    \put(0,0){\includegraphics[width=\unitlength,page=3]{repair_train.pdf}}%
    \put(0.01564194,0.16313547){\color[rgb]{0,0,0}\makebox(0,0)[lt]{\lineheight{1.25}\smash{\begin{tabular}[t]{l}$Y$\end{tabular}}}}%
    \put(0.92963519,0.16313547){\color[rgb]{0,0,0}\makebox(0,0)[lt]{\lineheight{1.25}\smash{\begin{tabular}[t]{l}$Y$\end{tabular}}}}%
    \put(0.41981089,0.16313547){\color[rgb]{0,0,0}\makebox(0,0)[lt]{\lineheight{1.25}\smash{\begin{tabular}[t]{l}RTT($Y$)\end{tabular}}}}%
    \put(0,0){\includegraphics[width=\unitlength,page=4]{repair_train.pdf}}%
  \end{picture}%
\endgroup%

    \caption{Training procedure of our APE model with RTT in language $Y$.}
    \label{fig:train_ape_model}
\end{figure}

We formalize our APE model as a translation model from synthetic ``translationese" \cite{gellerstam1986translationese} text in one language to natural text in the same language. For a language pair ($X$, $Y$) and a monolingual corpus $M_Y$ in language $Y$, the training procedure is as follows:
\begin{enumerate}
	\item Train two translation models on bitext for $X{\to}Y$ and $Y{\to}X$
	\item Use these models to generate round-trip translations for every target-language sentence $y$ in $M_Y$, resulting in the synthetic dataset $\textrm{RTT}(M_Y)$.
	\item Train a translation model on pairs of $(\textrm{RTT}(y), y)$, that translates from the round-tripped version of a sentence to its original form.
\end{enumerate}
This procedure is illustrated in Figure~\ref{fig:train_ape_model}.

\subsection{Application}
\begin{figure}[ht]
    \centering
    \def\svgwidth{\columnwidth}
\begingroup%
  \makeatletter%
  \providecommand\color[2][]{%
    \errmessage{(Inkscape) Color is used for the text in Inkscape, but the package 'color.sty' is not loaded}%
    \renewcommand\color[2][]{}%
  }%
  \providecommand\transparent[1]{%
    \errmessage{(Inkscape) Transparency is used (non-zero) for the text in Inkscape, but the package 'transparent.sty' is not loaded}%
    \renewcommand\transparent[1]{}%
  }%
  \providecommand\rotatebox[2]{#2}%
  \newcommand*\fsize{\dimexpr\f@size pt\relax}%
  \newcommand*\lineheight[1]{\fontsize{\fsize}{#1\fsize}\selectfont}%
  \ifx\svgwidth\undefined%
    \setlength{\unitlength}{204.79209271bp}%
    \ifx\svgscale\undefined%
      \relax%
    \else%
      \setlength{\unitlength}{\unitlength * \real{\svgscale}}%
    \fi%
  \else%
    \setlength{\unitlength}{\svgwidth}%
  \fi%
  \global\let\svgwidth\undefined%
  \global\let\svgscale\undefined%
  \makeatother%
  \begin{picture}(1,0.29163064)%
    \lineheight{1}%
    \setlength\tabcolsep{0pt}%
    \put(0,0){\includegraphics[width=\unitlength,page=1]{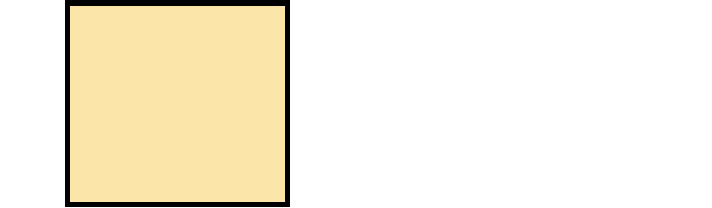}}%
    \put(0.12709192,0.15925224){\color[rgb]{0,0,0}\makebox(0,0)[lt]{\lineheight{1.25}\smash{\begin{tabular}[t]{l}\ \ \ \ \ NMT\\\ \ \ \ $X$$\to$$Y$\end{tabular}}}}%
    \put(0,0){\includegraphics[width=\unitlength,page=2]{repair_infer.pdf}}%
    \put(0.68009226,0.15925224){\color[rgb]{0,0,0}\makebox(0,0)[lt]{\lineheight{1.25}\smash{\begin{tabular}[t]{l}\ APE\\Model\end{tabular}}}}%
    \put(0,0){\includegraphics[width=\unitlength,page=3]{repair_infer.pdf}}%
    \put(0.00831743,0.15581096){\color[rgb]{0,0,0}\makebox(0,0)[lt]{\lineheight{1.25}\smash{\begin{tabular}[t]{l}$X$\end{tabular}}}}%
    \put(0.92963519,0.15581096){\color[rgb]{0,0,0}\makebox(0,0)[lt]{\lineheight{1.25}\smash{\begin{tabular}[t]{l}$Y$\end{tabular}}}}%
    \put(0.47108245,0.15581096){\color[rgb]{0,0,0}\makebox(0,0)[lt]{\lineheight{1.25}\smash{\begin{tabular}[t]{l}$Y$\end{tabular}}}}%
    \put(0,0){\includegraphics[width=\unitlength,page=4]{repair_infer.pdf}}%
  \end{picture}%
\endgroup%

    \caption{Automatic Post-Editing (APE) as post-processor of NMT.}
    \label{fig:apply_ape_model}
\end{figure}

Given a trained translation model and a trained APE model, the procedure is simply to a) translate any source text from language $X$ to language $Y$ with the translation model, and b) post-edit the output of the translation by passing it through the APE model. In this sense, the APE model may also be viewed as a paraphrasing model to produce ``naturalized" text. This procedure is illustrated in Figure~\ref{fig:apply_ape_model}.

\section{Experimental Setup}
\subsection{Architecture}
For the translation models, we use the transformer implementation in {\it lingvo}~\cite{shen2019lingvo}, using the transformer-base model size for Romanian$\to$English and transformer-big model size~\cite{vaswani2017attention} for German$\to$English and French$\to$English. The reverse models, English$\to$Romanian, English$\to$German and English$\to$French, are all transformer-big. All use a vocabulary of 32k subword units, and exponentially moving averaging of checkpoints (EMA decay) with the weight decrease parameter set to $\alpha=0.999$~\cite{buduma2017fundamentals}.

The APE models are also transformer models with 32k subword units and EMA decay trained with lingvo. For the German and the French APE models, we use the transformer-big size, whereas for the Romanian APE model, we use the smaller transformer-base setup as we have less monolingual data.

\subsection{Evaluation}
\label{subsec:eval}
We report \BLEU \cite{papineni2002bleu} and human evaluations. All \BLEU scores are calculated with sacreBLEU \cite{post2018call}\footnote{sacreBLEU signatures: BLEU+case.mixed+lang.en-LANG+numrefs.1+smooth.exp+SET+tok.intl+version.1.2.20}. 

Since 2014, the organizers of the WMT evaluation campaign \cite{bojar2017findings} have created test sets with the following method:
first, they crawled monolingual data in both English and the target language from news stories from online sources.
Thereafter they took about 1500 English sentences and translated them into the target language, and an additional 1500 sentences from
 the target language and translated them into English. This results in test sets of about 3000
sentences for each English-X language pair. The sgm files of each WMT test set include the original language for each sentence.

Therefore, in addition to reporting overall \BLEU scores on the different test sets, we also report results on the two subsets (based on the original language) of each newstest20XX, which we call the \{German,French,Romanian\}-original and English-original halves of the test set.
This is motivated by \newcite{Koppel:2011:TD:2002472.2002636}, who demonstrated that they can train a simple classifier to distinguish human-translated text from natural text with high accuracy. 
These text categorization experiments suggest that both the source language and the mere fact of being translated play a crucial role in the makeup of a
translated text. One of the major goals of our APE model is to rephrase the NMT output in a more natural way, aiming to remove undesirable translation artifacts that have been introduced.

To collect human rankings, we present each output to crowd-workers, who were asked to score each
sentence on a 5-point scale for:
\begin{itemize}
    \item \textbf{fluency:} How do you judge the overall naturalness of the utterance in terms of its grammatical
correctness and fluency?
\end{itemize}
Further, we included the source sentence and asked the raters to evaluate each sentence on a 2-point scale (binary decision) for:
\begin{itemize}
    \item \textbf{accuracy:} Does the statement factually contradict anything in the reference information?
\end{itemize}

Each task was given to three different raters. Consequently, each output has a separate score for each question that is the average of 3 different ratings. 

\subsection{Data}
For the round-trip experiments we use the monolingual News Crawl data from the WMT evaluation campaign. We remove duplicates and apply a max-length filter on the source sentences and the round-trip translations, filtering to the minimum of 500 characters or 70 tokens. For German, we concatenate all News Crawl data from 2007 to 2017, comprising 216.5M sentences after filtering and removing duplicates. For Romanian, we use News Crawl '16, comprising 2.2M sentences after filtering and deduplication. For French, we concatenate News Crawl data from 2007 to 2014, comprising 34M sentences after filtering.

Our translation models are trained on WMT18 ($\sim$5M sentences for German after filtering), WMT16 ($\sim$0.5M sentences for Romanian after filtering) and WMT15 ($\sim$41M sentences for French) bitext. For Romanian and German we filter sentence pairs that have empty source or target, that have source or target longer than 250 tokens, or the ratio of whose length is greater than 2.0. 
For English$\to$German and English$\to$French, we also build a system based on noised back-translation, as in \newcite{edunov2018understanding}. We use the same monolingual sentences that we used for the APE model to generate the noisy back-translation data.

\section{Experiments}

\subsection{English$\to$German}

\begin{table*}[t!]
\begin{center}
    {\setlength{\tabcolsep}{.3em}
    \begin{tabular}{|l|c|c|c|c|c|}
        \hline
         & \multicolumn{1}{|c}{\bf{newstest2014}} & \multicolumn{1}{|c}{\bf{newstest2015}} & \multicolumn{1}{|c}{\bf{newstest2016}} & \multicolumn{1}{|c}{\bf{newstest2017}} & \multicolumn{1}{|c|}{\bf{average}} \\ \hline
                    \newcite{vaswani2017attention} & 28.4 & - & - & - & \\ \hline
                    \newcite{shaw2018self} & 29.2 & - & - & - & \\ \hline
                    our bitext & 29.2 & 31.4 & 35.0 & 29.4 & 31.2 \\ \hline
                    \ + RTT APE & 30.7 & 31.2 & 33.6 & 30.1 & 31.4\\ \hline
                    \ + RTT APE de-orig only & 31.7 & 32.9 & 37.2 & 31.9 & 33.4\\ \hline \hline
                    our NBT & 33.5 & 34.4 & 38.3 & 32.5 & 34.7 \\ \hline
                     \ + RTT APE  & 32.5 & 32.7 & 35.2 & 31.3 & 32.9 \\ \hline
                     \ + RTT APE de-orig only  & 34.0 & 34.5 & 38.7 & 33.2 & 35.1\\ \hline
    \end{tabular}
    \caption{\BLEU scores for WMT18 English$\to$German. We apply the same APE model (trained on RTT with bitext models) for both an NMT system based on pure bitext and an NMT system that uses noised back-translation (NBT) in addition to bitext.}
    \label{tab:result-wmt18-ende}
    }
    \end{center}
\end{table*}

\begin{table*}[th!]
\begin{center}
    {\setlength{\tabcolsep}{.3em}
    \begin{tabular}{|l|c|c|c|c|c|c|c|c|c|c|}
        \hline
        & \multicolumn{2}{|c}{\bf{newstest2014}} & \multicolumn{2}{|c}{\bf{newstest2015}} & \multicolumn{2}{|c}{\bf{newstest2016}} & \multicolumn{2}{|c}{\bf{newstest2017}} & \multicolumn{2}{|c|}{\bf{average}} \\
                    & orig-de & orig-en & orig-de & orig-en  & orig-de & orig-en & orig-de &orig-en & orig-de & orig-en\\ \hline \hline
                    our bitext & 28.4 & 29.4 & 26.5 & 33.3 & 29.9 & 38.2 & 25.9 & 31.6 & 27.7 & 33.1 \\ \hline
                    \ + RTT APE & 34.1 & 27.6 & 31.3 & 30.9 & 35.7 & 32.2 & 32.1 & 28.5 & 33.3 & 29.8 \\ \hline \hline
                    our NBT & 35.6 & 31.3 & 32.6 & 34.7 & 37.6 & 38.7 & 31.7 & 32.6 & 34.4 & 34.3 \\ \hline
                    \ + RTT APE & 36.9 & 28.8 & 33.5 & 32.0 & 38.5 & 32.9 & 33.8 & 29.2 & 35.7 & 30.7 \\ \hline
    \end{tabular}
    \caption{\BLEU scores for WMT18 English$\to$German. Test sets are divided by their original source language (either German or English).}
    \label{tab:result-wmt18-ende-origlan}
    }
    \end{center}
\end{table*}

\begin{table*}[th!]
    \centering
    \begin{tabular}{|l|c|c|c|c|c|c|c|c|}
        \hline
        & \multicolumn{4}{|c|}{\bf{newstest2016}} & \multicolumn{4}{|c|}{\bf{newstest2017}}\\
        & \multicolumn{2}{|c}{fluency} & \multicolumn{2}{|c|}{accuracy} & \multicolumn{2}{|c}{fluency} & \multicolumn{2}{|c|}{accuracy} \\ \hline
        & orig-de & orig-en & orig-de & orig-en & orig-de & orig-en & orig-de & orig-en\\ \hline \hline
        baseline bitext & 4.65 & 4.49 & 95.6\% & 94.4\% & 4.74 & 4.52 & 97.2\% & 94.6\% \\ \hline
        \ + RTT APE & 4.77 & 4.59 & 98.4\% & 95.0\% & 4.84 & 4.58 & 98.0\% & 94.8\%\\ \hline \hline
        our NBT & 4.79 & 4.64 & 98.2\% & 95.8\% & 4.79 & 4.65 & 98.2\% & 95.4\%\\ \hline
        \ + RTT APE & 4.82 & 4.63 & 98.0\% & 96.2\% & 4.86 & 4.68 & 98.0\% & 96.4\% \\ \hline \hline
        reference & 4.85 & 4.67 & 98.6\% & 98.6\% & 4.83 & 4.70 & 98.0\% & 99.2\% \\ \hline
    \end{tabular}
    \caption{English$\to$German human evaluation results split by original language of the test set.}
    \label{tab:human_results-ende}
\end{table*}

The results of our English$\to$German experiments are shown in Table~\ref{tab:result-wmt18-ende}. We trained the APE model on RTT produced by English$\to$German and German$\to$English NMT models that are only trained on bitext. Applying the APE model on the output of our NMT model also trained on only bitext  improves the \BLEU scores by up to 1.5 \BLEU points for newstest2014 and 0.7 \BLEU points for newstest2017. Nevertheless, the score drops by 1.4 points on newstest2016. To investigate the differing impact on the test sets, we split each test set by its original language (Table~\ref{tab:result-wmt18-ende-origlan}). The APE model consistently increases the \BLEU on the German-original half of the test set, but decreases the \BLEU on the English-original half. Consequentially, we applied our APE model only on the sentences with original language in German (\emph{+RTT APE de-orig only} in Table \ref{tab:result-wmt18-ende}) and see consistent improvements over all test sets with an average \BLEU improvement of 2.2 points.

To verify that the drop in \BLEU score is because of the unnatural reference translations, we run a human evaluation (see Section~\ref{subsec:eval}) for both fluency/grammatical correctness and accuracy. Based on the human ratings (Table~\ref{tab:human_results-ende}), our APE model also improves on the English-original half of the test set (which is a more realistic use case).

\ 

Without re-training, we use the APE model that is trained on the bitext RTT and apply it to a stronger NMT system that also includes all the available monolingual data in the form of noised back-translation. We see a very similar pattern to the previous experiments. Regarding automatic scores, our APE model only improves on the German-original part of the test sets, with an average improvement of 1.3 \BLEU points. The human evaluations show the same inconsistency with the automatic scores for the English-original half. As with the weaker baseline, humans rate the output of our APE model at least as fluent and accurate as the original output of the NMT model (Table~\ref{tab:human_results-ende}). Further, we also run a human evaluation on the reference sentences and found that the scores for both fluency and accuracy are only minimally higher than for our APE NBT output.

Comparing only the \BLEU scores from our bitext and NBT models in Table~\ref{tab:result-wmt18-ende-origlan} reveals that augmenting the parallel data with back-translated data also mostly improves the \BLEU scores on the German-original half of the test set. This is in line with the results of our APE model and opens the question of how much of the original bitext data is natural on the target side.

As our APE model seems agnostic to the model which produced the RTT, we applied it to the best submissions of the recent WMT18 evaluation campaign, applying to German-original half of the test set only. Table~\ref{tab:result-wmt18-campaign-ende} shows the results for the 2 top submissions of Microsoft \cite{junczysdowmunt:2018:WMT18} and Cambridge \cite{stahlberg-degispert-byrne:2018:WMT}. Both systems improved by up to 0.8 points in \BLEU.

\begin{table}[ht]
    \centering
    \begin{tabular}{|l|c|c|}
        \hline
        & Microsoft & Cambridge  \\ \hline \hline
         WMT18 submission & 48.7 & 47.2\\ \hline
         \ + APE only de-orig & 49.5 & 47.7 \\ \hline
    \end{tabular}
    \caption{\BLEU scores for WMT18 English$\to$German newstest2018. We apply our APE model only on the German-original half of the test set. \BLEU scores are calculated on the full newstest2018 set and the English-original half is just copied from the submission.}
    \label{tab:result-wmt18-campaign-ende}
\end{table}

\begin{figure}[ht]
    \includegraphics[width=\linewidth]{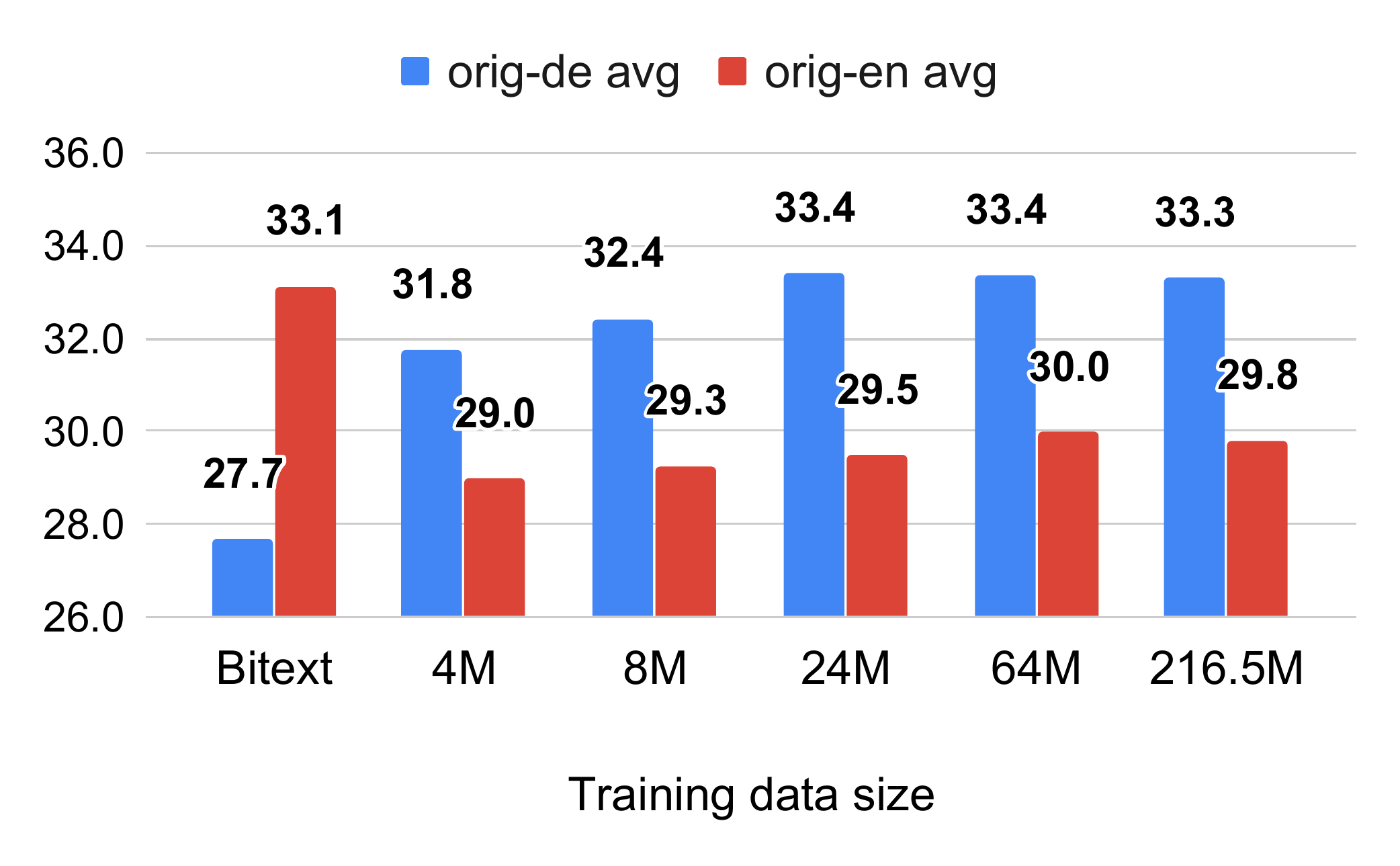}
    \caption{English$\to$German - Average \BLEU scores for newstest2014-newstest2017: Our APE model is trained on different subsets of the monolingual data.}
    \label{fig:subsets-ende}
\end{figure}

Finally, we train our APE model on different random subsets of the available 216.5M monolingual data (see Figure~\ref{fig:subsets-ende}).
The average \BLEU scores on newstest2014-newstest2017 show that we can achieve similar performance by using 24 million training examples only, and that large improvements are seen using as few as 4M training examples. 

\subsection{English$\to$Romanian}
Experimental results for the WMT16 English$\to$Romanian task are summarized in Table~\ref{tab:result-wmt18-enro}. By applying our APE model on top of a
baseline that is only trained on bitext, we see improvements of 3.0 \BLEU (dev) and 0.3 \BLEU (test) over our baseline system when we automatically post edit only to the Romanian-original half of the test set.
Similar to English$\to$German, we apply our APE model on the top 2 submissions of the WMT16 evaluation campaign (Table~\ref{tab:result-wmt16-campaign-enro}).
Both the QT21 submission \cite{peter-EtAl:2016:WMT16}, which is a system combination of several NMT systems, and the ensemble
of the University of Edinburgh \cite{sennrich-haddow-birch:2016:WMT} improve, by 0.3 \BLEU and 0.2 \BLEU on test, respectively.

\begin{table}[ht]
\begin{center}
    {\setlength{\tabcolsep}{.3em}
    \begin{tabular}{|l|c|c|c|c|}
        \hline
         & \multicolumn{1}{|c}{\bf{dev}} & \multicolumn{1}{|c|}{\bf{test}} \\ \hline
                    \newcite{sennrich-haddow-birch:2016:WMT} & - & 28.8 \\ \hline
                    our bitext & 27.0 & 28.9  \\ \hline
                    \ + RTT APE & 27.3 & 29.0 \\ \hline
                    \ + RTT APE only ro-orig & 30.0 & 29.2 \\ \hline
    \end{tabular}
    \caption{\BLEU scores for our models on WMT16 English$\to$Romanian.}
    \label{tab:result-wmt18-enro}
    }
    \end{center}
\end{table}

\begin{table}[ht]
    \centering
    \begin{tabular}{|l|c|c|}
        \hline
        & QT21 & Edinburgh  \\ \hline \hline
         WMT16 submission & 29.4 & 28.8 \\ \hline
         \ + RTT APE only ro-orig & 29.7 & 29.0 \\ \hline
    \end{tabular}
    \caption{\BLEU scores for WMT16 English$\to$Romanian test set. Our APE model was applied on top of the best WMT16 submissions.}
    \label{tab:result-wmt16-campaign-enro}
\end{table}

\subsection{English$\to$French}

Experimental results for English$\to$French are summarized in Table~\ref{tab:result-wmt15-enfr}. We see the same tendency as we saw for German and Romanian. When applying our APE system on the output of the bitext baseline, we get a small improvement of 0.1 \BLEU. By only post-editing the French-original half, we get an improvement of 1.0 \BLEU points. The same effect can be seen on the English$\to$French system that is trained with Noised BT. We yield quality improvements of 0.8 \BLEU by applying our APE model on the French-original half of the test set only.

\begin{table}[ht]
\begin{center}
    {\setlength{\tabcolsep}{.3em}
    \begin{tabular}{|l|c|c|}
        \hline
         & \multicolumn{1}{|c|}{\bf{newstest2014}} \\ \hline
                    our bitext & 43.2 \\ \hline
                    \ + RTT APE & 43.3 \\ \hline
                    \ + RTT APE only fr-orig & 44.2 \\ \hline \hline
                    our NBT & 45.3 \\ \hline
                    \ + RTT APE & 44.6 \\ \hline
                    \ + RTT APE only fr-orig & 46.1 \\ \hline
    \end{tabular}
    \caption{\BLEU scores for WMT15 English$\to$French.}
    \label{tab:result-wmt15-enfr}
    }
    \end{center}
\end{table}

\section{Example Output}
We would like to highlight a few short examples where our APE model improves the NMT translation in German.
Although our APE model is also quite helpful for long sentences, we will focus on short examples for the sake of simplicity.
In Table~\ref{tab:example_output_ape} there are examples from the English$\to$German noised back-translated (NBT) setup (see Table~\ref{tab:result-wmt18-ende}),
with and without automatic post editing.
In the first example, NMT translates \emph{club} (i.e. cudgel) incorrectly into \emph{Club} (i.e. organization). Based on the context of the sentence, our APE model learned that
\emph{club} has to be translated into \emph{Schlagstock} (i.e. cudgel). 
The next two examples are very similar as our APE model improves the word choice of the translations by
taking the context of the sentence into account. 
The NMT translations of the last two examples make little sense and our APE model rephrases the output into a fluent, meaningful sentence.

\begin{table*}[ht]
    \centering
    \setlength\tabcolsep{4pt}
    \begin{tabular}{c|l}
    \hline
    source & Using a \textbf{club}, they \textbf{beat} the victim in the face and upper leg. \\
    NBT & Mit einem \textbf{Club schlagen} sie das Opfer in Gesicht und Oberschenkel. \\
    \ + RTT APE & Mit einem \textbf{Schlagstock schlugen} sie dem Opfer ins Gesicht und in den Oberschenkel. \\ \hline
    source & M\"uller put another one in with a \textbf{with a penalty}. \\
    NBT & M\"uller setzte einen weiteren \textbf{mit einer Strafe} ein. \\
    \ + RTT APE & M\"uller netzte einen weiteren \textbf{per Elfmeter} ein. \\ \hline
    source & Obama \textbf{receives} Netanyahu \\
    NBT & Obama \textbf{erh\"alt} Netanjahu \\
    \ + RTT APE & Obama \textbf{empf\"angt} Netanjahu \\ \hline
    source & At least one Bayern fan was \textbf{taken injured from the stadium}. \\
    NBT & Mindestens ein Bayern-Fan wurde \textbf{vom Stadion verletzt}. \\
    \ + RTT APE & Mindestens ein Bayern-Fan wurde \textbf{verletzt aus dem Stadion gebracht}. \\ \hline
    source & The archaeologists \textbf{made a find in the third construction phase} of the Rhein Boulevard. \\
    NBT & Die Arch\"aologen \textbf{haben in der dritten Bauphase} des Rheinboulevards \textbf{gefunden}. \\
    \ + RTT APE & Die Arch\"aologen \textbf{sind im dritten Bauabschnitt} des Rheinboulevards \textbf{f\"undig geworden}. \\ \hline
    \end{tabular}
    \caption{Example output for English$\to$German.}
\label{tab:example_output_ape}
\end{table*}

\section{Discussion}
In this section, we focus on the results on target-language-original test sets, like the English-original subset of newstest2016 (Table~\ref{tab:result-wmt18-ende-origlan} and Table~\ref{tab:human_results-ende}), where the APE model lowered the score by 6 \BLEU, yet improved human evaluations. A na{\" i}ve take-away from this result would be that evaluation sets whose target side is natural text are inherently superior. However, translating \textit{from} translationese also has its own problems, including 1) it does not represent any real-world translation task, and 2) translationese sources may be much easier to translate ``correctly", and reward MT biases like word-for-word translation. 
The take-away, therefore, must be to report scores both on the source-language-original and the target-language-original test sets, rather than lumping two test-sets together into one as has heretofore been done. This gives a higher-precision glimpse into the strengths and weaknesses of different modeling techniques, and may prevent some effects (like improvements in naturalness of output) from being hidden.

Going forward, our results should also be seen as a call for higher-quality test sets.
Multi reference \BLEU is one option and less likely to suffer these biases as acutely, and has previously been used in the NIST projects. Another option could be to align sentence pairs from monolingual data sets in two languages and run human evaluation to exclude bad sentence pairs.

\section{Related Work}

\textbf{Automatic Post-Editing} \\
Probably most similar to our work, \newcite{junczys2016log,junczys2018ms} uses RTT as additional training data for the automatic post-editing (APE) task of the WMT evaluation campaign \cite{chatterjee2018findings}. They claimed that the provided post-editing data is orders of magnitude too small to train neural models, and combined the training data with artificial training data generated with RTT. They found that the additional artificial data helps against early overfitting and makes it possible to overcome the problem of too little training data. 
In contrast to our work, they do not report results for models only trained on the artificial RTT data. Further, their RTT data is much smaller (10M sentences) compared to ours (up to 200M sentences) and they only report results for the APE subtask.

There have been several earlier approaches using RTT for APE. \newcite{hermet2009using} used RTT to improve a standard preposition error detection system. Although
their evaluation corpus was limited to 133 prepositions, the hybrid system outperformed their standard method by roughly 13\%.
\newcite{madnani2012exploring} combined RTT obtained from Google Translate via 8 different pivot languages into a lattice for grammatical error correction. Similar to system combination, their final output is extracted by the shortest path scored by different features. They claimed that their preliminary experiments yield fairly satisfactory results but leave significant room for improvement.

\textbf{Back-translation} \\
Back-translation \cite{sennrich2016improving,poncelas2018investigating} augments relatively scarce parallel data with plentiful monolingual data, allowing one to train source-to-target (S2T) models with the help of target-to-source (T2S) models.
Specifically, given a set of sentences in the target language, a pre-constructed T2S translation system is used to generate translations to the source language. These synthetic sentence pairs are combined with the original bilingual data when training the S2T NMT model.

\textbf{Iterative Back-translation} \\
Iterative back-translation \cite{zhang2018joint, cotterell2018explaining, hoang2018iterative} is a joint training algorithm to enhance the effect of monolingual source and target data by iteratively boosting the source-to-target and target-to-source translation models. The joint training method uses the monolingual data and updates NMT models through several iterations. A variety of flavors of iterative back-translation have been proposed, including \newcite{niu2018bidirectional}, who simultaneously perform iterative S2T and T2S back-translation in a multilingual model, and  \newcite{he2016dual, xia2017dual}, who combine dual learning with phases of back- and forward-translation. 

\newcite{artetxe2018unsupervised,artetxe2018unsupervisedb} and \newcite{lample2017unsupervised,lample2018phrase} used iterative back-translation to train two unsupervised translation systems in both directions ($X{\to}Y$ and $Y{\to}X$) in parallel. Further, they used back-translation to generate a synthetic source to construct a dev set for tuning the parameters of their unsupervised statistical machine translation system. In a similar formulation, \newcite{cheng2016semisupervised} jointly learn a translation system with a round-trip autoencoder.

\textbf{Round-tripping and Paraphrasing} \\
Round-trip translation has seen success as a method to generate paraphrases.
\newcite{bannard2005paraphrasing} extracted paraphrases by using alternative phase translations from bilingual phrase tables from Statistical Machine Translation.
\newcite{mallinson2017paraphrasing} presented PARANET, a neural
paraphrasing model based on round-trip translations with NMT. They showed that their paraphrase model outperforms all traditional paraphrase models.

\newcite{wu2018improving} train a paraphrasing model on $(X, \textrm{RTT}(X))$ pairs, translating from natural text into a simplified version. They apply this sentence-simplifier on the source sentences of the training data and report quality gains for IWSLT. 

\textbf{Translationese and Artifacts from NMT} \\
The difference between translated sentence pairs based on whether the source or the target is the original sentence has long been recognized by the human translation community, but only partially investigated by the machine translation community. An introduction to the latter is presented in \newcite{Koppel:2011:TD:2002472.2002636}, who train a high-accuracy classifier to distinguish human-translated text from natural text in the Europarl corpus. This is in line with research from the professional translation community, which has seen various works investigating both systematic biases inherent to translated texts \cite{baker1993corpus, selinker1972interlanguage}, as well as biases resulting specifically from interference from the source text \cite{toury1995descriptive}. There has similarly long been a focus on the conflict between \textit{Fidelity} (the extent to which the translation is faithful to the source) and \textit{Transparency} (the extent to which the translation appears to be a natural sentence in the target language) \cite{warner2018the, schleiermacher1816ueber, dryden1685preface}. To frame our hypotheses in these terms, the present work hypothesizes that outputs from NMT systems often err on the side of disfluent fidelity, or word-by-word translation.

There are a few papers that discuss the effect of translationese on MT models. \newcite{lembersky2012language, stymne2017effect} explored how the translation direction
for statistical machine translation affects the translation result. They found that using
training and tuning data translated in the same direction as the translation systems tends to give the best results.
\newcite{holmqvist2009improving} noted that the original language of the test sentences influences the \BLEU score of translations. They showed that the \BLEU scores for target-original sentences are on average higher than sentences that have their original source in a different language.
\newcite{popel2018cuni} split the WMT Czech-English test set based on the original language. They found that when training on synthetic data, the model performs much better on the Czech-original half than on the non Czech-original half. When trained on authentic data, it is the other way round.
\newcite{fomicheva2017role} found that both the average score and Pearson correlation with human judgments is substantially higher when both the MT output and human translation were generated from the same source language.

\section{Ablation}

\subsection{Iterative APE}
We can apply our APE model in an iterative fashion several times on the same output. In Table~\ref{tab:result-wmt18-ende-origlan-iterative}, we applied our APE model on the already post-edited output to see if we can further naturalize the sentences. As a result, 75\% of the sentences did not change. The remaining sentences lowered the \BLEU scores on average by 0.1 points for German-original half and by 0.7 points for the English-original half of the test sets.

\begin{table}[ht]
\begin{center}
    {\setlength{\tabcolsep}{.3em}
    \begin{tabular}{|l|c|c|}
        \hline
 & \multicolumn{2}{|c|}{\bf{average}} \\
                    & orig-de & orig-en\\ \hline \hline
                    our bitext & 27.7 & 33.1 \\ \hline
                    \ + APE & 33.3 & 29.8 \\ \hline
                    \ + 2xAPE & 33.2 & 29.1 \\ \hline
    \end{tabular}
    \caption{Average \BLEU scores for WMT18 English$\to$German newstest2014-2017. We run our APE model a second time on the output of the already post-editied output.}
    \label{tab:result-wmt18-ende-origlan-iterative}
    }
    \end{center}
\end{table}

\subsection{Reverse APE}
Instead of training an APE model on $(\textrm{RTT}(y), y)$ sentence pairs (see Section~\ref{sec:ape_model}), we train in this section a \texttt{reverse APE} model that flips source and target and is trained on $(y, \textrm{RTT}(y))$ sentence pairs. Experimental results can be seen in Table~\ref{tab:result-wmt18-ende-reverse}. Overall, the performance decreases on both the German-original half and the English-original half. Interestingly, the \BLEU scores of the reverse APE model on the English-original half are higher than those of the normal APE model. This demonstrates again that sentences evaluated on the English-original half prefer output that is biased by the translation process.

\begin{table}[ht]
\begin{center}
    {\setlength{\tabcolsep}{.3em}
    \begin{tabular}{|l|c|c|}
        \hline
        & \multicolumn{2}{|c|}{\bf{average}} \\
                    & orig-de & orig-en\\ \hline \hline
                    our bitext & 27.7 & 33.1 \\ \hline
                    \ + RTT APE & 33.3 & 29.8 \\ \hline
                    \ + Reverse APE & 25.1 & 30.6 \\ \hline \hline
                    our NBT & 34.4 & 34.3 \\ \hline
                    \ + RTT APE & 35.7 & 30.7 \\ \hline
                    \ + Reverse APE & 27.0 & 31.3 \\ \hline
    \end{tabular}
    \caption{Average \BLEU scores for WMT18 English$\to$German newstest2014-2017.}
    \label{tab:result-wmt18-ende-reverse}
    }
    \end{center}
\end{table}

\subsection{Inside the black box of RTT}
In this subsection we are interested in how much RTT changes translation outputs.
We calculate the \BLEU scores of all English$\to$German test sets (11,175 sentences in total) by taking the original German sentences as references and their RTT as hypotheses.
Although the RTT hypotheses are a  less clean (paraphrased) version of the references, having been forward-translated from an already noisy back-translated source, the \BLEU score is 40.9, with unigram precision of 72.3\%, bigram precision of 48.9\%, trigram precision of 35.6\% and 4gram precision of 26.6\%.

We further found that the original sentences use a larger vocabulary than the artificial RTT data. While the output of the RTT has only 29,635 unique tokens, the original sentences contain 33,814 unique tokens. Even more interesting, the NMT output (from the model trained only on bitext) of the same test sets has a vocabulary size of 30,540, but after running our APE on the same test sets the vocabulary size increases to 31,471. The NMT output from the NBT model has a vocabulary size of 32,170 and its post-edited version increases the number of unique words to 32,283. Overall, we see that both the RTT and the NMT output have a smaller vocabulary size than the original or post-edited versions, and that \BLEU score grows directly with increased number of unique tokens in the target side.

\section{Conclusion}
We propose an APE model that is only trained on RTT and increases the quality of NMT translations, measured both by \BLEU and human evaluation. We see improvements both when automatically post editing our  model translations and when automatically post editing outputs from the winning submissions to the WMT competition. Our APE has the advantage that it is agnostic to the model which produced the translations, and so can be used on top of any new advance in the field, without need for re-training. Further, we demonstrate that we need only a subset of 24M training examples to train our APE model. We furthermore use this model as a tool to reveal systematic problems with reference translations, and propose finer-grained \BLEU reporting on both source-language-original test sets and target-language-original test sets, as well as calling for higher-quality and multi-reference test sets.

\bibliography{acl2018}
\bibliographystyle{acl_natbib}

\end{document}